\documentclass[aps,preprint,showpacs,superscriptaddress]{revtex4-1}  % for double-spaced preprint
\usepackage{graphicx}
\usepackage{subcaption}
\usepackage{bm}   
\usepackage{color}
\usepackage{amssymb}
\usepackage{amsmath}
\usepackage{float}
\usepackage{lineno,hyperref}
\usepackage{graphicx}
\usepackage{dcolumn}
\usepackage{bm}
\usepackage{enumitem}
\usepackage[utf8]{inputenc}
\usepackage[T1]{fontenc}
\usepackage{mathptmx}
\usepackage{mathtools}
\usepackage{amsmath}
\usepackage{rotating}
\usepackage{adjustbox}
\modulolinenumbers[5]
\begin{document}
	\title{Prediction of Occurrence of Extreme Events using Machine Learning}
	\author{J. Meiyazhagan}
	\affiliation{Department of Nonlinear Dynamics, Bharathidasan University, Tiruchirappalli - 620 024, Tamilnadu, India}
	\author{S. Sudharsan}
	\affiliation{Department of Nonlinear Dynamics, Bharathidasan University, Tiruchirappalli - 620 024, Tamilnadu, India}
	\author{A. Venkatesan}
	\affiliation{PG and Research Department of Physics, Nehru Memorial College (Autonomous),  Puthanampatti, Tiruchirappalli 621 007, Tamil Nadu, India.}
	\author{M. Senthilvelan}
	\email[Correspondence to: ]{velan@cnld.bdu.ac.in}
	\affiliation{Department of Nonlinear Dynamics, Bharathidasan University, Tiruchirappalli - 620 024, Tamilnadu, India}
	\vspace{10pt}
	
	\begin{abstract}
	Machine learning models play a vital role in the prediction task in several fields of study. In this work, we utilize the ability of machine learning algorithms to predict the occurrence of extreme events in a nonlinear mechanical system. Extreme events are rare events that occur ubiquitously in nature. We consider four machine learning models, namely Logistic Regression, Support Vector Machine, Random Forest and Multi-Layer Perceptron in our prediction task. We train these four machine learning models using training set data and compute the performance of each model using the test set data. We show that the Multi-Layer Perceptron model performs better among the four models in the prediction of extreme events in the considered system. The persistent behaviour of the considered machine learning models is cross-checked with randomly shuffled training set and test set data.
	\end{abstract}
	
	%\maketitle
	\maketitle
\section{Introduction}
Rare events that are ubiquitously observed in physical, biological, societal and environmental systems are called extreme events and they are well known for their destructive consequences. Floods, cyclones, tsunamis, tornadoes, earthquakes, droughts, epidemics, epileptic seizures and rogue waves are some of the well-known examples of extreme events \cite{dysthe20081,jentsch2005,krause20151}. The devastating aftermath effects of these events are severe to the society. So it is important to carry out multifaceted research on extreme events and such studies will aid us in creating tools to reduce the impact that they produce. In the study of extreme events, among several perspectives, focus has been mainly made on the following four important aspects, namely (i) prediction, (ii) mechanism, (iii) mitigation and (iv) statistics \cite{fara2008}.  Prediction involves the determination of reliable indicators or measurable observables of extreme events. Mechanism deals with the conditions and precursors that involve in the occurrence of extreme events. Mitigation involves the formulation of strategies to control extreme events. Statistics involves extreme value theory and large deviation theory in order to quantify the extreme events from statistical point of view. In recent years, studies on extreme events are mainly focused on determining the mechanism behind the occurrence of extreme events in dynamical systems such as Li\'enard \cite{kingston2017extreme}, microelectro mechanical system, solid $\textrm{CO}_2$ laser \cite{kumarasamy2018extreme}, loss-modulated $\textrm{CO}_2$ laser \cite{bonatto2017extreme} and El Ni\~no southern oscillation \cite{ray2020understanding}. For further detailed review on extreme events, one can refer the review articles in Refs. \cite{fara2008,chowdhury2021extreme}. Although the mechanism has been analyzed thoroughly from the dynamical systems point of view, even in simple systems, the prediction of these events is yet to be precisely determined. However, it is the most important task that has to be understood to avoid the catastrophic effects. As far as the prediction task is concerned, it helps to forecast well in advance when and where the extreme events can occur. In general, prediction determines the value of some observables at a specific time in future from the available data of the past. Although few works have been made on prediction, all those classical prediction methods give reliable results only when the system is simple and it turns out to be a difficult task when the system is complex or chaotic. To overcome the failure of classical predictability, recently, Machine Learning (ML) methods have been invoked for prediction purposes.

\par In recent years, ML has become a predominant tool in prediction tasks in various fields of physics \cite{Lohani2020,Carleo2019,Radovic2018,sudhe1,sudhe2,sudhe3,santo1}. The term $``$Machine Learning$"$ was coined long ago \cite{AS} and ML is defined as a field of study that gives computers the ability to learn without being explicitly programmed. Prediction tasks that are generally used in ML are regression, classification, clustering, dimensionality reduction and reinforcement learning \cite{goodfellow2016}. Among these, classification and regression are often been used in many fields of study. In classification, the algorithm attempts to classify the input data among the preset categories whereas in regression the algorithm predicts the {numerical value(s) in }output by learning {the relationship between given outputs and} various input features \cite{goodfellow2016}. 

\par Nowadays the ML models are prevalently being used in the study of nonlinear dynamical systems in order to predict various dynamical states. In particular, ML models have been used to identify the chimera states \cite{BARMPARIS2020,ganaie2020,kushwaha2020}, replicate chaotic attractors \cite{Pathak2017}, predict the amplitude of chaotic laser pulses \cite{Amil2019}, detect unstable periodic orbits \cite{zhu2019}, to separate chaotic signals \cite{Krishnagopal2020} and network classification from symbolic time-series \cite{Panday2021}. In particular, in the study of extreme events a well known ML technique, namely reservoir computing has been used to  predict the extreme events in coupled FitzHugh-Nagumo model \cite{PYRAGAS2020}, whereas another ML algorithm called Artificial Neural Network (ANN) has been used to predict extreme events in the H\'enon map \cite{lellep2020}. On the other hand, wavelet-based ML methods have been employed to forecast extreme flood events in the flood-prone river basin \cite{Yeditha2020}. In addition to this, deep learning methods have also been used as a forecasting tool in the study of extreme events \cite{meiysudha1,dibak1,ray2021optimized,asch2021}. We note here that in the above works, only regression type methodology was adopted in order to predict extreme events except the work done in Ref.~\cite{lellep2020} in which the authors have considered the prediction task as a classification problem and they took the points on the state space of the H\'enon attractor as input of the ANN and the output was classified as either extreme or non-extreme. 
%--------------------------------------------------------------------------
\par In the present work, we consider a non-polynomial mechanical system that describes the dynamics of a particle in a rotating parabola. From the application point of view, this mechanical model describes a motorbike being ridden in a rotating parabolic well in a circus, centrifugation devices, centrifugal filters and industrial hoppers \cite{Venkatesan1997,centrifugation,filter,hopper}. Extreme events in these systems may represent a sudden increase in the velocity of the particle inside the rotating system which may cause sudden damage to the system. Predicting extreme events in this system well in advance will definitely help us to avoid unexpected accidents to the particle and also to the system itself. With the help of ML models, we aim to classify the values of the system's parameters into extreme and non-extreme. The system parameters which we consider are (i) external forcing strength, (ii) strength of the parametric drive and (iii) strength of time-delay feedback. The respective values of the parameters for which the system exhibits (does not exhibit) extreme events are classified as extreme (non-extreme). We intend to carry out this classification task using four ML models, namely (i) Logistic Regression , (ii) Support Vector Machine , (iii) Random Forest \cite{chandramouli2018machine} and (iv) Multi-Layer Perceptron \cite{goodfellow2016}. Initially, the data are randomly shuffled before beginning the training process in particular 450 different values of the parameters are fed as input to the ML model, trained and 150 values of the parameters are used for prediction purposes. A similar task is performed using four more random shuffles. In all the five cases, it is found that all the four models successfully perform the prediction task and the prediction accuracy of all the models are also highly impressive. Further, in order to determine the best model among the four, the performance of these models are also captured. For the purpose of determining the performance of these models, first, we compute the confusion matrix which has a total number of True Positives, True Negatives, False Positives and False Negatives of the prediction result. From the elements of the confusion matrix, we determine the following metrics, that are (i) accuracy, (ii) precision, (iii) recall and (iv) F1 score. Our results indicate that MLP performs better than the other three considered models in the task of prediction of extreme events. We also observe that the performance given by MLP is immensely consistent even in the case of random input data. To the best of our knowledge, this is the first time the values of the parameters are classified into occurrence and non-occurrence of extreme events using ML models.  

\par We formulate the paper in the following way. In Sec. 2, we describe a  non-polynomial mechanical system and discuss the occurrence of extreme events in that system. In Sec. 3, we discuss the steps involved in generating input and output data for the training and testing of the ML models and characteristics of that generated data. In Sec. 4, we consider four different ML models and train them with the training set data. To choose the best model among the four for the prediction of extreme events, we analyze the performance of each model by calculating the performance metrics in Sec. 5.  Finally, we present our conclusion in Sec. 6.

\begin{figure}
	\centering
	\includegraphics[width=1.0\linewidth]{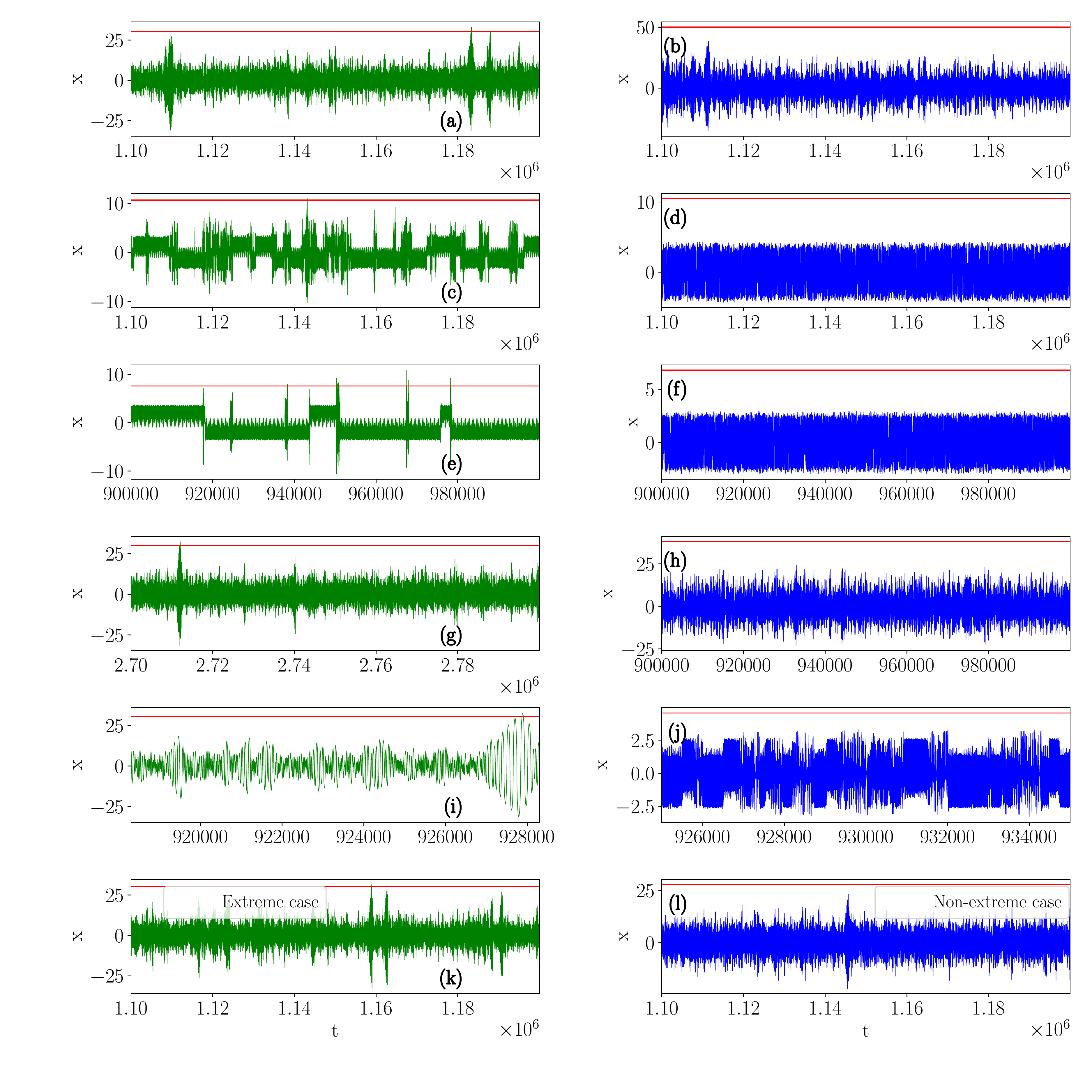}
	\caption{\label{fig:data_ts} Example trajectories of the system \eqref{gen} in both extreme and non-extreme cases for the six combinations of parameter values. The green colour and blue colour denoting the occurrence and absence of extreme events respectively. The parameters are taken as (a) $f = 0, \epsilon=0.081, \delta=5\times 10^{-4}$, (b) $f = 0, \epsilon=0.1, \delta=5\times 10^{-4}$, (c) $f = 3.055, \epsilon=0, \delta=0.019$, (d) $f = 3.055, \epsilon=0, \delta=0.04$, (e) $f = 3.1665, \epsilon=0, \delta=0$, (f) $f = 1.8, \epsilon=0, \delta=0$, (g) $f = 1.0, \epsilon=0.081, \delta=0$, (h) $f = 1.8, \epsilon=0.081, \delta=0$, (i) $f = 0, \epsilon=0.081, \delta=0$, (j) $f = 0, \epsilon=0.05, \delta=0$, (k) $f = 0.001, \epsilon=0.081, \delta=5\times 10^{-4}$ and (l) $f = 0.1, \epsilon=0.081, \delta=5\times 10^{-4}$. All other parameters in Eq.~(\ref{gen}) are taken commonly as $\lambda=0.5$, $\omega_0^2 = 0.25$, $\omega_p=1.0$, $\alpha = 0.2$, $\tau=0.1$ and $\omega_e = 1.0$}
\end{figure}
\section{Extreme events in non-polynomial mechanical system}
\begin{figure}[!ht]
	\centering
	\includegraphics[width=0.5\linewidth]{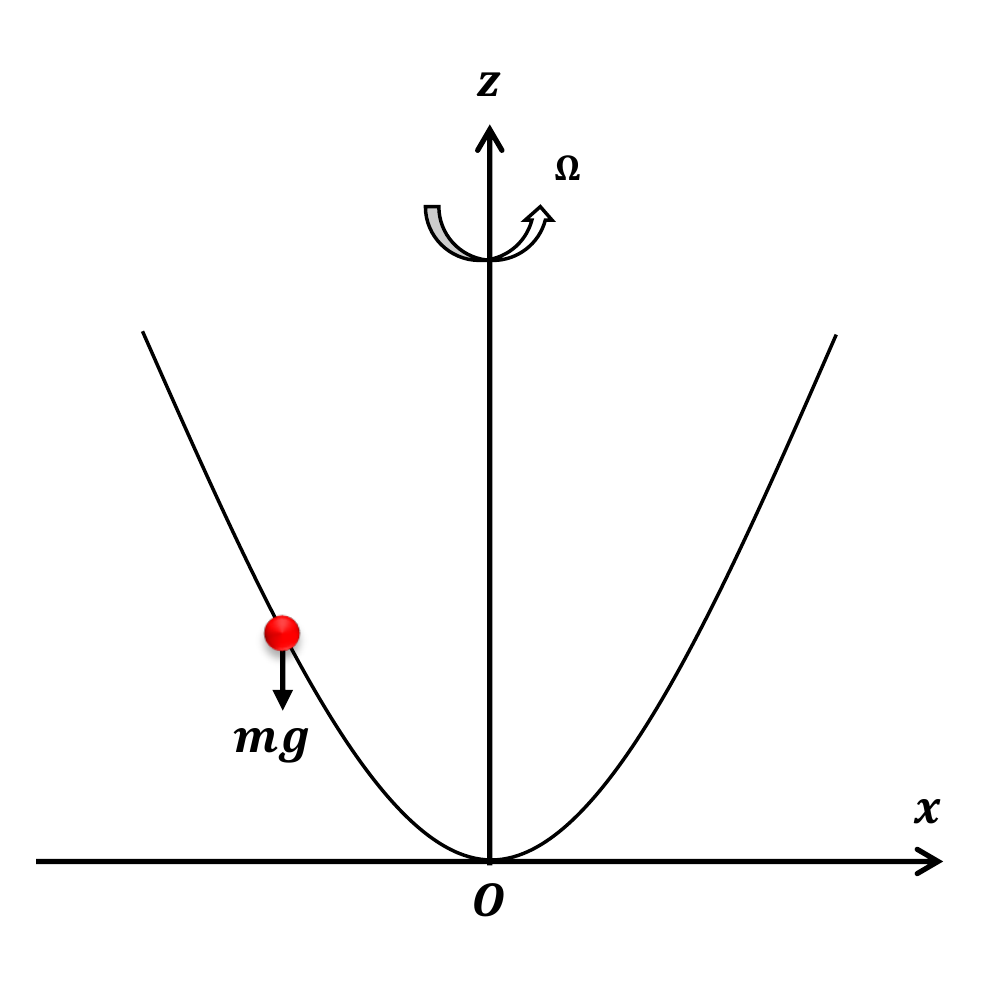}
	\caption{\label{fig:system} Representation of the considered dynamical system. The red colour solid circle denotes the motion of a freely sliding particle on a parabolic wire rotating with constant angular velocity $\Omega$}
\end{figure}

We consider a mechanical model describing the motion of a freely sliding particle of unit mass on a parabolic wire ($z = \sqrt{\lambda} x^2$) rotating with a constant angular velocity $\Omega$ ($\Omega^2 = \Omega_0^2 = -\omega_0^2 + g\sqrt{\lambda}$) as shown in Fig.~\ref{fig:system}. 
Here $g$ is the acceleration due to gravity, $1/\sqrt{\lambda}$ is the semi-latus rectum of the rotating parabola, and $\omega_0$ is the initial angular velocity. The equation describing the aforesaid dynamics with additional damping, external periodic forcing, parametrically driven angular velocity and time-delayed feedback can be modelled by the following second order nonlinear differential equation \cite{Venkatesan1997,suresh2018}, namely
\begin{equation}
\begin{aligned}[b]
(1+\lambda x^2)\ddot{x}&+\lambda x\dot{x}^2+\omega_0^2x-\Omega_0^2[2\epsilon~\mathrm{cos}~\omega_pt\\  &+0.5\epsilon^2(1+\mathrm{cos}~2\omega_pt)]x+\alpha\dot{x} + \delta x_\tau =f~\mathrm{cos}~\omega_e t. \label{gen}
\end{aligned}
\end{equation}

Here overdot denotes differentiation with respect to $t$, $\alpha$ is the damping parameter, $f$ and $\omega_e$ are respectively the strength and frequency of the external periodic forcing. Further, $\omega_p$ and $\epsilon$ are the frequency and the strength of the parametric drive and $\delta$ is the strength of the time delay feedback, while $x_\tau=x(t-\tau)$ provides the time delay $\tau$. In one of our recent works \cite{ss1}, we have shown that extreme events appear under the influence of parametric drive ($\epsilon \neq 0$, $f=\delta=0$) and those events were mitigated when external forcing $f$ was introduced along with $\epsilon$.

\par  Although the emergence of extreme events was reported and the mechanism behind the emergence was determined in a sub-case, prediction of extreme events, which is the most important task, has yet to be carried out in the system \eqref{gen}. So in the present work, for the first time, we predict the occurrence of extreme events accurately in the system \eqref{gen} as a classification task involving three parameters, namely $f$, $\epsilon$ and $\delta$ through four ML models. The importance of this study lies in the fact that this prediction task does not include any detailed dynamical analysis to determine extreme events, but the trained model accurately tells whether the extreme events have occurred or not just by inputting the values of system parameters to the models. An added advantage of this study is that once the model is trained for a range of parameters for which the dynamics is known, it will accurately predict even for the range of parameter values for which the dynamics is unknown. Further, the ML models that we have adopted in this work to predict extreme events will help us to devise safety measures to protect the system and the particles in the system from the damage caused by extreme events.

In the system \eqref{gen}, an event is said to be extreme if the trajectory of $x$ or $\dot{x}$ crosses the qualifier threshold value. This threshold value is calculated by the formula \cite{dysthe20081}
\begin{equation}
x_{ee}=\langle x\rangle+4\sigma_x,
\end{equation}
where $\langle x\rangle$ is the mean peak amplitude and $\sigma_x$ is the standard deviation of peak values.

\section{Data Generation}
\par For any kind of prediction using ML, the first and foremost task is to generate the data. Our foremost aim is to train the model in order to predict whether, for a given set of values of parameters, the system exhibits an extreme event or not. To do this, we fix the parameters in \eqref{gen} as $\lambda=0.5$, $\omega_0^2 = 0.25$, $\omega_p=1.0$, $\alpha = 0.2$, $\tau=0.1$ and $\omega_e = 1.0$. The other three parameters, that is $\epsilon$, $f$ and $\delta$, are the bifurcation parameters and so we feed them as input to the considered ML model. For the input we consider the following six combinations:
%---------------------------------------------------------------------

%------------------------------------------------------------------------

\begin{enumerate}[label=(\alph*)]
	\item \textit{Time delay with parametric drive ($f = 0, \epsilon \neq 0, \delta \neq 0$)} 
	\par {In this case, we consider the system {\eqref{gen}} with parametric drive and time delay feedback in the absence of external forcing. The system with this setup may or may not produce extreme events which depends on the values of the considered parameters. One such example is shown in the Figs.~{\ref{fig:data_ts}} (a) \& (b). In Fig.~{\ref{fig:data_ts}} (a) we can see that the trajectories of the system crosses the threshold value $x_{ee}$ for the values $f = 0, \epsilon=0.081$ and $\delta=5\times 10^{-4}$ confirming the occurrence of extreme events while from Fig.~{\ref{fig:data_ts}} (b), we can infer that for the parameter values $f = 0, \epsilon=0.1$ and  $\delta=5\times 10^{-4}$ there is no emergence of extreme events. Hence as far as the input data, in this case, is concerned, the value of the external forcing parameter can be considered as zero and the other two parameters have non-zero values.}
	
	\item \textit{Time delay with forcing ($f \neq 0, \epsilon=0, \delta \neq 0$)}
	\par {Now we set the system {\eqref{gen}} with external forcing and time delay feedback in the absence of parametric drive. As in the previous case, here also the system in the considered setup will produce extreme events for a set of parameter values and no extreme events for another set of values. We have shown an example for both extreme and non-extreme cases in Figs.~{\ref{fig:data_ts}} (c) and (d) respectively. The trajectory of the system corresponding to the values $f = 3.055, \epsilon=0$ and $\delta=0.019$ crosses the $x_{ee}$ (see Fig.~{\ref{fig:data_ts}} (c)) confirms that the system will produce extreme events with this setup. On examining the Fig.~{\ref{fig:data_ts}} (d), one may observe that there is no such events for the parameter values of $f = 3.055, \epsilon=0$ and $\delta=0.04$. The input data, in this case, will be non-zero for external forcing and time delay feedback while it is zero for a parametric drive.}
	
	\item \textit{External force only ($f \neq 0, \epsilon=0, \delta=0$)}
	\par {In this case, we consider our system {\eqref{gen}} with external forcing only. The system is now free from time delay feedback and the parametric drive. With this setup also we obtain extreme events as shown in the Fig.~{\ref{fig:data_ts}} (e) for the parameter values $f = 3.1665, \epsilon=0$ and $\delta=0$. For other set of parameter values, say for instance $f = 1.8, \epsilon=0$ and $\delta=0$ the system does not produce any extreme events which can be seen from Fig.~{\ref{fig:data_ts}} (f). When considering the input for this case, the time delay feedback and parametric drive values are taken as zero and for external forcing value as non-zero.}
	
	\item \textit{Parametric drive with forcing ($f \neq 0, \epsilon\neq 0, \delta=0$)} 
	\par {Here we consider the system with parametric drive and external forcing but without time delay feedback. In this case, we are able to generate another set of data that corresponds to both extreme and non-extreme cases. We illustrate the system behaviour with this setup for two sets of parameter values in Figs.~{\ref{fig:data_ts}} (g) and (h). Figure~{\ref{fig:data_ts}} (g) is an example for the extreme case. The corresponding parameter values are $f = 1.0, \epsilon=0.081$ and $\delta=0$. Figure~{\ref{fig:data_ts}} (h) shows the non-extreme case corresponding to the parameter values $f = 1.8, \epsilon=0.081$ and $\delta=0$. For this case, the input parameter values are taken as zero for time delay feedback and non-zero values for parametric drive and forcing.}
	
	\item \textit{Parametric drive only ($f=0, \epsilon \neq 0, \delta=0$)}
	\par {In this case, the system acts under the influence of parametric drive only and the other two parameters, namely external forcing strength and time delay feedback strength are not taken into account. The system with this setup also produces extreme events for certain parameter values and no extreme events for another set of parameter values. We present an example for both extreme and non-extreme cases in Figs.~{\ref{fig:data_ts}} (i) and (j). In Fig.~{\ref{fig:data_ts}} (i), the time series is plotted which in turn shows the occurrence of extreme events for the parameter values $f = 0, \epsilon=0.081$ and $\delta=0$ and Fig.~{\ref{fig:data_ts}} (j) is plotted for the non-extreme case for the parameter values $f = 0, \epsilon=0.05$ and $\delta=0$. In the present case, we take non-zero values for the parametric drive and zero for both the external forcing and time delay feedback in both extreme and non-extreme regimes.}
	
	\item \textit{Time delay with parametric drive and forcing ($f \neq 0, \epsilon \neq 0, \delta \neq 0$)}
	\par {As a last case, we consider the system {\eqref{gen}} in the presence of all three parameters, namely strength of external forcing, parametric drive and time delay feedback strength. The system with this setup also exhibits extreme events for specific set of values. We show that the occurrence of extreme events in the above system in Fig.~{\ref{fig:data_ts}} (k) for the parameter values $f = 0.001, \epsilon=0.081$ and $\delta=5\times 10^{-4}$ and the Fig.~{\ref{fig:data_ts}} (l) represents the non-occurrence of extreme events for the parameter values $f = 0.1, \epsilon=0.081$ and $\delta=5\times 10^{-4}$. As far as the input data is concerned all the three parameters are taken as non-zero values.}
\end{enumerate}
Since the problem that we consider is a binary classification problem, we classify all the above mentioned six combinations of parameter values into extreme and non-extreme events by taking them as input for the ML models.

\begin{table}[!ht]
	\centering
	\begin{tabular}{|c|c|c|c|c|c|c|c|c|}
		\hline
		\multicolumn{9}{|c|}{Distribution of data}\\\hline 
		&\multicolumn{8}{|c|}{Train set}\\ \hline
		No.  & No. of extreme & No. of non-extreme & (a) & (b) & (c) & (d) & (e) & (f) \\ \hline
		1 & 230  & 220  & 74 & 76 & 74 & 73 & 76 & 77 \\
		\hline
		2 & 233  & 217  & 70 & 77 & 79 & 72 & 78 & 74 \\
		\hline
		3 & 228  & 222 &  71 & 78 & 79 & 69 & 69 & 84 \\
		\hline
		4 & 223  & 227 &  68 & 74 & 73 & 81 & 80 & 74 \\
		\hline
		5 & 220  & 230 &  76 & 80 & 75 & 78 & 74 & 67 \\
		\hline\hline\hline
		&\multicolumn{8}{|c|}{Test set}\\ \hline
		No. & No. of extreme & No. of non-extreme & (a) & (b) & (c) & (d) & (e) & (f) \\ \hline
		1 &  70 & 80 & 29 & 23 & 26 & 29 & 21 & 22\\
		\hline
		2  & 67 & 83 & 30 & 23 & 21 & 28 & 22 & 26\\
		\hline
		3  & 72 & 78 & 29 & 22 & 21 & 31 & 31 & 16\\
		\hline
		4 & 77 & 73 & 32 & 26 & 27 & 19 & 20 & 26\\
		\hline
		5 & 80 & 70 & 24 & 20 & 25 & 22 & 26 & 33\\
		\hline
	\end{tabular}
	\caption{Count of classes in train and test set data for five randomly shuffled data. Here the counts in (a) - (f) denote the total number of data in six combinations in generated data}
	\label{tab:distri}
\end{table}
%--------------------------------------------------------

The time series trajectories shown in the left panel of Fig.~\ref{fig:data_ts} correspond to the parameter values for which the system exhibits extreme events. From the time series, the occurrence of extreme events is confirmed if the trajectories of the time series cross a threshold value $x_{ee}$. The threshold value is represented by a horizontal line in Fig.~\ref{fig:data_ts}. The time series trajectories which are shown in the right panel correspond to the parameter values for which the system does not exhibit extreme events. We generate 50 extreme and 50 non-extreme cases for all six combinations of the parameters. On the whole, we collect 600 values, out of which 300 values come out from extreme events emerging regime and the other 300 values represent regimes where no extreme events are seen. We feed the parameter values of $f,\epsilon$ and $\delta$ as input vector \textbf{X} to the ML models and in the output vector \textbf{Y}, we classify extreme cases as 1 and non-extreme cases as 0. To start with, we randomly shuffle the data and split the data into two sets, namely the training set and the testing set.  Training is the process of forcing the ML models to find and optimize the relation between the output with the given input data and testing is the process of predicting the output for a given input and checking whether the predicted output is matching with the actual output. With this objective, we randomly choose 450 data for training and 150 data for testing. The input data is scaled to its mean and variance before being feed into the ML models. We consider five randomly chosen data sets as mentioned in Table~\ref{tab:distri}. It is important to check the distribution of each class in both the training and testing set data. In Table~\ref{tab:distri}, one may notice that both in the training and testing set the number of extreme and non-extreme cases are more or less equal. While making the random shuffle, it is also equally important to check whether the number of values is equally distributed in each one of the six combinations ((a)-(f) as mentioned previously) of parameters both in the training and testing data. All these details are given in Table~\ref{tab:distri}. This type of distribution helps to avoid the priority given by the ML algorithm to a particular case (extreme or non-extreme) or to a particular combination of parameters while training the data. With these primary checks, the algorithm will work better on the test set data. {For example, during the training process the number of data set for the combination (a) is higher than that of the number of data in the combination (b), then the model may give more importance to the case (a) than to the case (b). When it comes to the testing phase, there may be some chances for the model which fail to predict the correct results for the case (b) because the model has been trained only for a few values on the case (b). So to avoid this we generate data in all six combinations more or less equally.}
%------------------------------------------------------

%--------------------------------------------------------
\section{Machine Learning models}
\par {All} the generated data {are now ready for the training of ML models,} namely (i) Logistic Regression, (ii) Support Vector Machine, (iii) Random Forest and (iv) Multi-Layer Perceptron which we consider in our work. {In this section, we present the mathematical description and working functionality of each model.}
\subsection{Logistic Regression (LR)}
\par
\begin{figure}[!ht]
	\centering
	\includegraphics[width=0.7\linewidth]{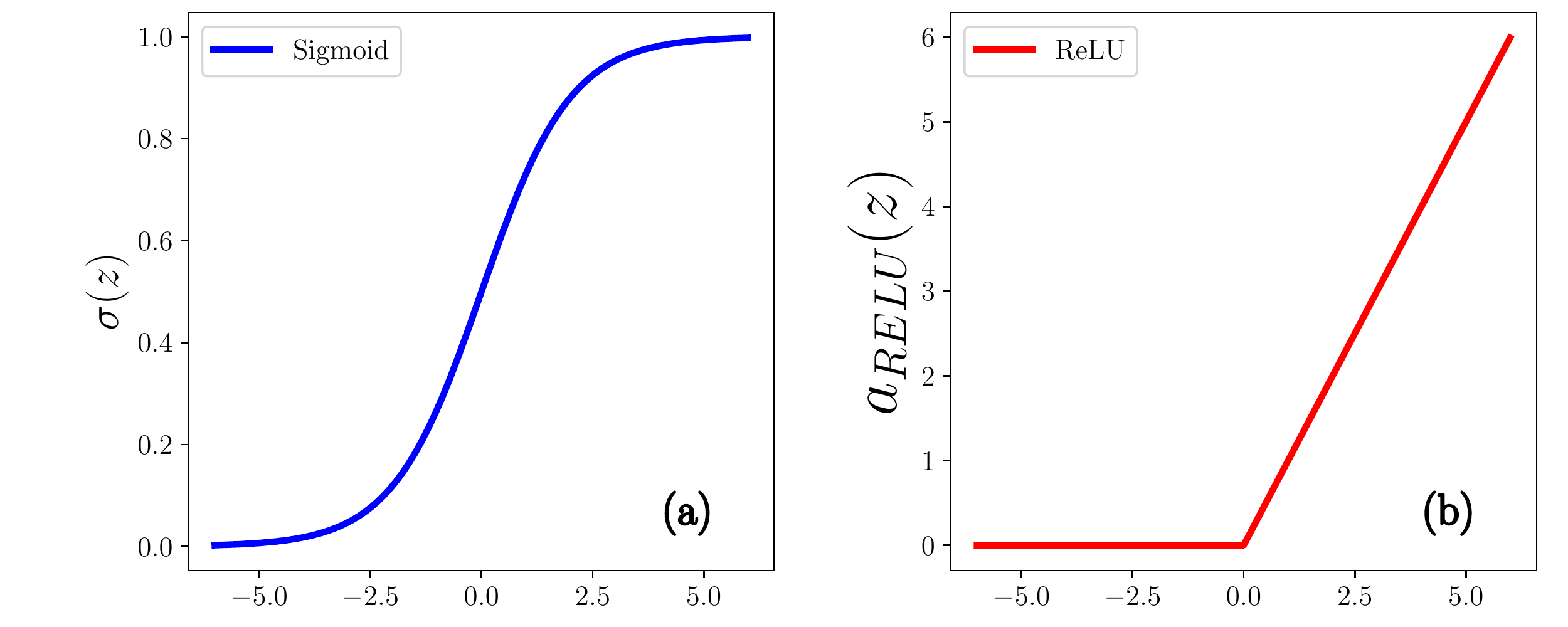}
	\caption{\label{fig:ac_fn} Activation functions used in the ML algorithms. (a) sigmoid activation function used in LR and output layer activation in MLP, (b) ReLU function used for the activations of hidden layers in MLP}
\end{figure}

 LR algorithm is a type of regression analysis often used in binary classification problems {which have outputs like  $``$0 or 1$"$, $``$high or low$"$, $``$yes or no$"$ and $``$true or false$"$}. {Since} it is a linear classifier, the output of this model depends on the linear function \cite{chandramouli2018machine}
\begin{equation}
f(X) = b_0 + b_1 x_1 + b_2 x_2 + \hdots +b_n x_n,
\end{equation}
where {$X$ is the input vector which has values of all features (in our case they are nothing but the three parameter values $f,\epsilon,\delta$) and } $b_n$'s are the predicted weights and the value of $n$ corresponds to the number of features that is being fed as input. {Since we have three features in the input,} in our case the value of $n$ is 3. The output of this LR model can be achieved by applying a sigmoid function (shown in Fig.~\ref{fig:ac_fn}.(a)) to the above linear function, that is
\begin{subequations}
	\begin{equation}
	\hat{Y} = \sigma(f(X)),
	\end{equation}
	where { $\hat{Y}$ is the output of the LR model and $\sigma$ is the sigmoid function which takes of the form}
	\begin{equation}
	\sigma(z) = \frac{1}{1+e^{-z}}.\label{sigmoid_fn}
	\end{equation}
\end{subequations}
The output value of the sigmoid activation is ranging from 0 to 1 denoting the probability of each class for the given input data. {If the value of the probability is above some critical value then the output is 1 and if the value is below the critical value then the output is 0. The commonly used critical value is $0.5$}. {While concerning our case the final output is fixed }based on the value of probabilities of each class as 0 and 1 representing non-extreme and extreme cases respectively. 

.
%--------------------------------------------------------
\subsection{Support Vector Machine (SVM)}
\par SVM algorithm {is a discriminative classification algorithm suitable for the supervised learning problems. This model works on the principle of constructing a hyperplane that separates the given data into required classes. The hyperplane is constructed in the multi-dimensional feature space.} The dimension of the hyperplane is $(N-1)$ for a $N$-dimensional feature space. In our case, {we work with three features ($f,\epsilon,\delta$) and so we have }$3D$ feature space and {$2D$ hyperplane for the binary classification of extreme or non-extreme.} The data points near the hyperplane are called support vectors. SVM model uses the kernel method to transform a lower-dimensional input space into a higher dimensional space. In our work, we use Radial Basis Function (RBF) kernel, also known as Gaussian kernel, which is given by \cite{chandramouli2018machine} 
\begin{equation}
K(x_i,x_j)=exp\left( - \frac{\Vert{x_i-x_j}\Vert^2}{2S^2}\right),
\end{equation}
{where $\Vert{x_i-x_j}\Vert^2$  denotes the squared Euclidean distance between the points $x_i$ and $x_j$ and} $S$ is the standard deviation. {As mentioned above, after} the input data is fed, while training the model, a two dimensional hyperplane is generated on the three dimensional feature space { for the separation of given three parameter values into extreme and non-extreme cases. }

\begin{figure}[!ht]
	\centering
	\includegraphics[width=0.6\linewidth]{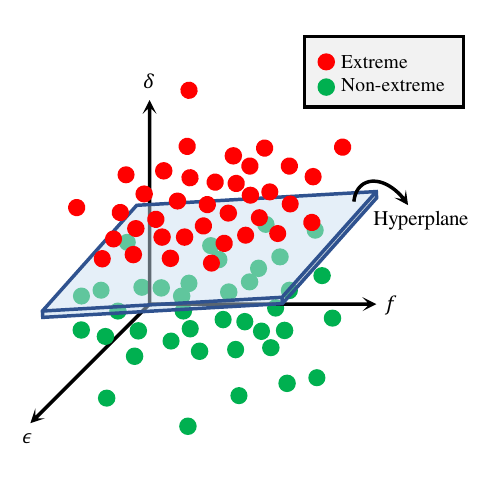}
	\caption{\label{fig:svm} Schematic diagram of 2D hyperplane in 3D feature space. The red and green circles are representing our two different classes of extreme and non-extreme respectively}
\end{figure}

{This constructed hyperplane acts as a boundary separating the two classes. Testing results of the model depends on this hyperplane. One can easily visualize the working of the SVM from the schematic diagram given in Fig.~{\ref{fig:svm}}. The solid circles represent the feature (parameter) values on the $3D$ space. Red and green circles, respectively represent the extreme and non-extreme cases. We can see that the $2D$ hyperplane in Fig.~{\ref{fig:svm}} separates the feature space into two. The space above the hyperplane corresponds to the extreme case and the space below the hyperplane corresponds to the non-extreme case. After training, if a new set of parameter values are given to the model then the result is based on the point's position (above or below) from the hyperplane. We note here that the representation of the model given in Fig.~{\ref{fig:svm}} is only a schematic diagram drawn for the purpose of understanding the hyperplane and feature space.}
%--------------------------------------------------------
\subsection{Random Forest (RF)}
\par RF \cite{Breiman} is an ensemble based classifier. The ensemble contains many decision trees. Ensembling is normally done using the concept of bagging with random feature subspaces \cite{chandramouli2018machine}. In our work, we use a large number of decision trees in RF in order to learn the contribution of each feature from the input data. In particular, we consider 1000 such trees in our study. To make classification on the given set of input data, the algorithm seeks to find which class is maximum in the outputs of the decision trees. This is performed by getting the outputs from each decision tree and the final output is selected by means of voting. In other words, the final output is either extreme or non-extreme which was the output of a maximum number of decision trees.
\begin{figure}[!ht]
	\centering
	\includegraphics[width=0.8\linewidth]{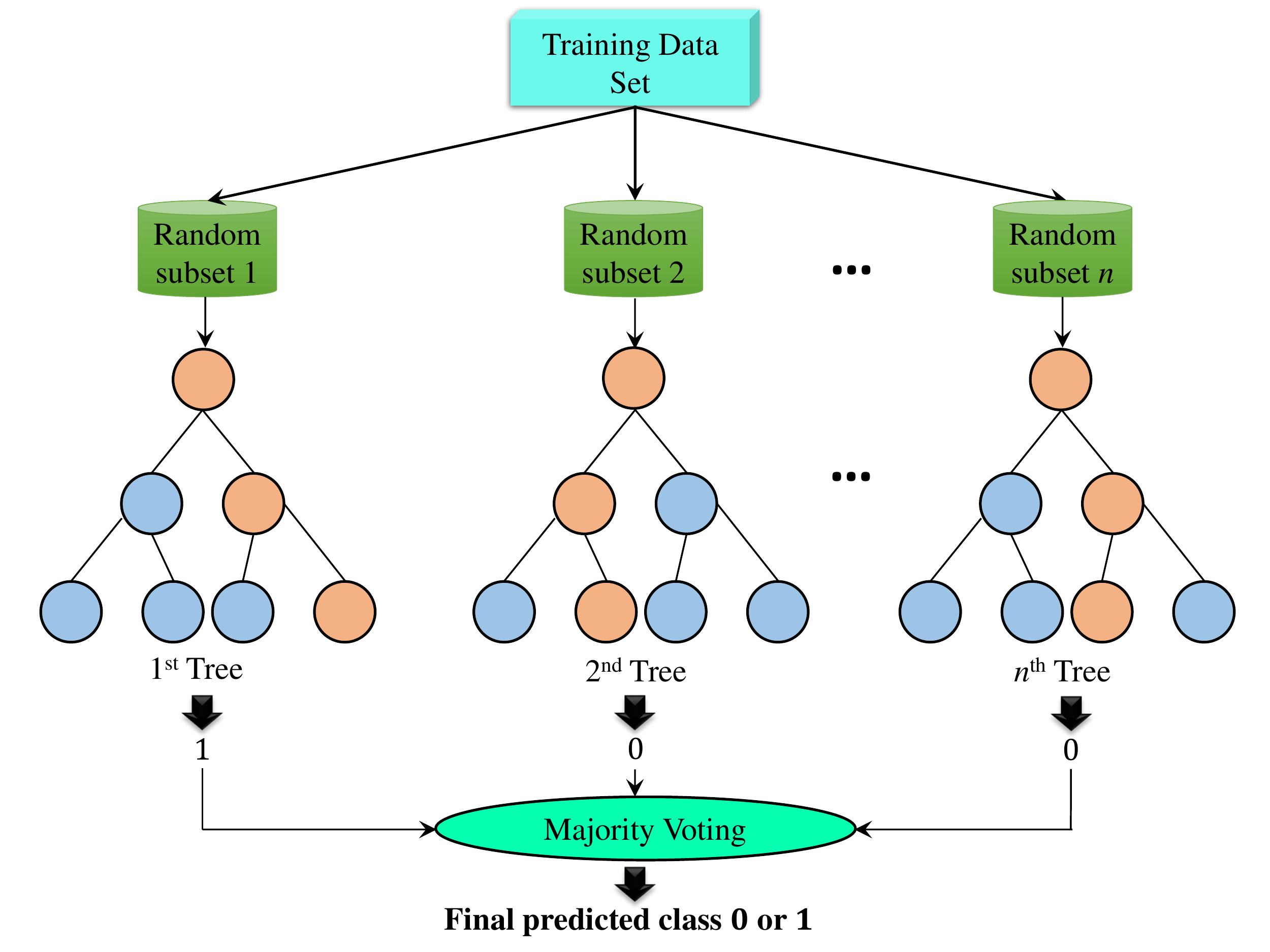}
	\caption{\label{fig:rf} Schematic diagram of RF}
\end{figure}
{We present the schematic working of the RF model in Fig.~{\ref{fig:rf}}. In this figure, we can see that the first step of the training process is the separation of the given training set data into random subset data. In our case, we have 1000 random subsets of data and each has a decision tree.  Each tree has branches in which the algorithm makes decisions on the classification task for the corresponding random set of data. After getting output from each decision tree the final decision is made upon voting. In other words, the class which has the majority outcomes from the trees will be the final output.}
%--------------------------------------------------------
\subsection{Multi-Layer Perceptron (MLP)}
\begin{figure}[!ht]
	\centering
	\includegraphics[width=1\linewidth]{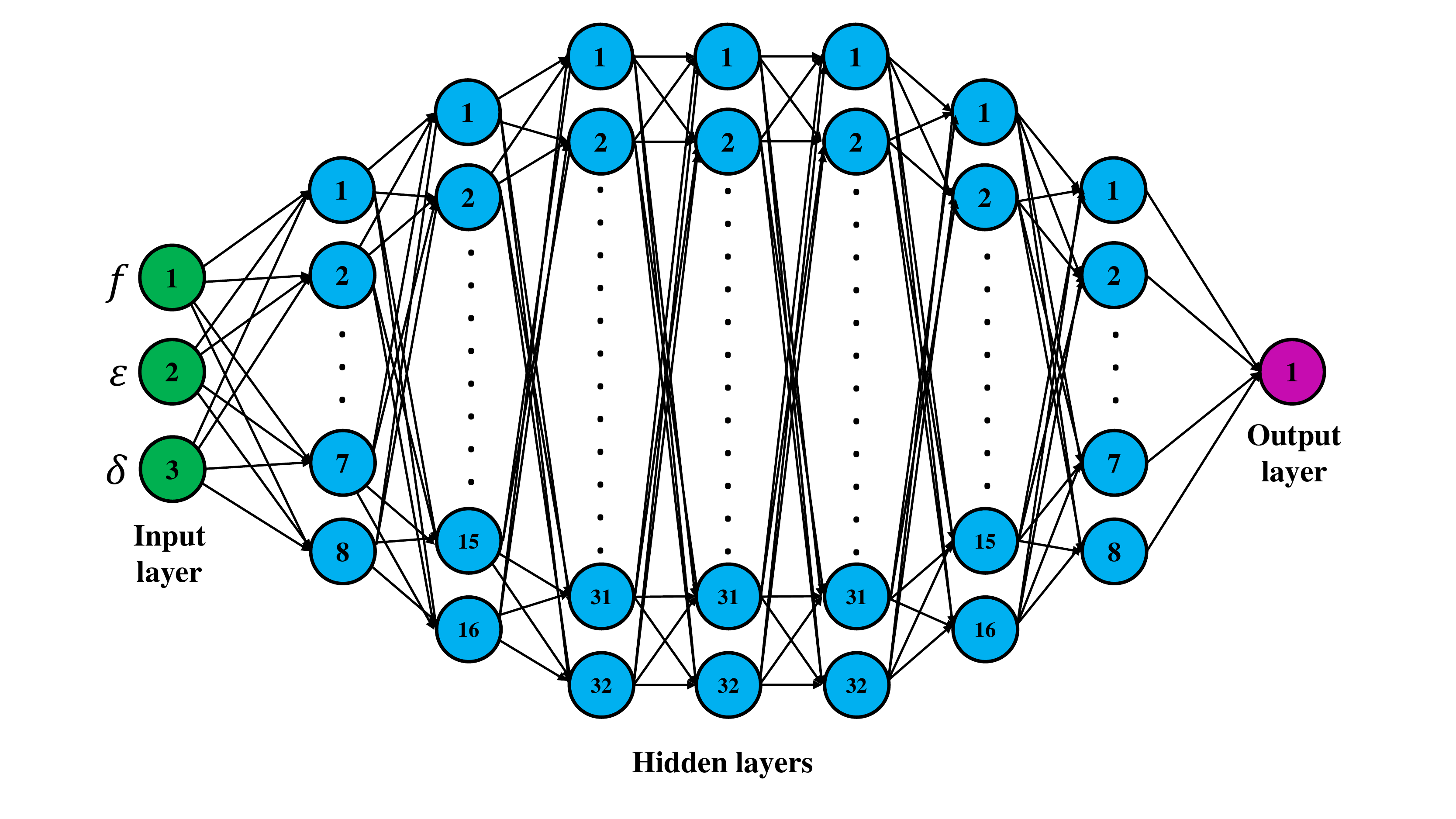}
	\caption{\label{fig:dnn} Schematic diagram of MLP with one input layer and one output layer. Between the input and output layer, there are seven hidden fully connected layers represented in blue colour}
\end{figure}

\begin{figure}[!ht]
	\centering
	\includegraphics[width=0.65\linewidth]{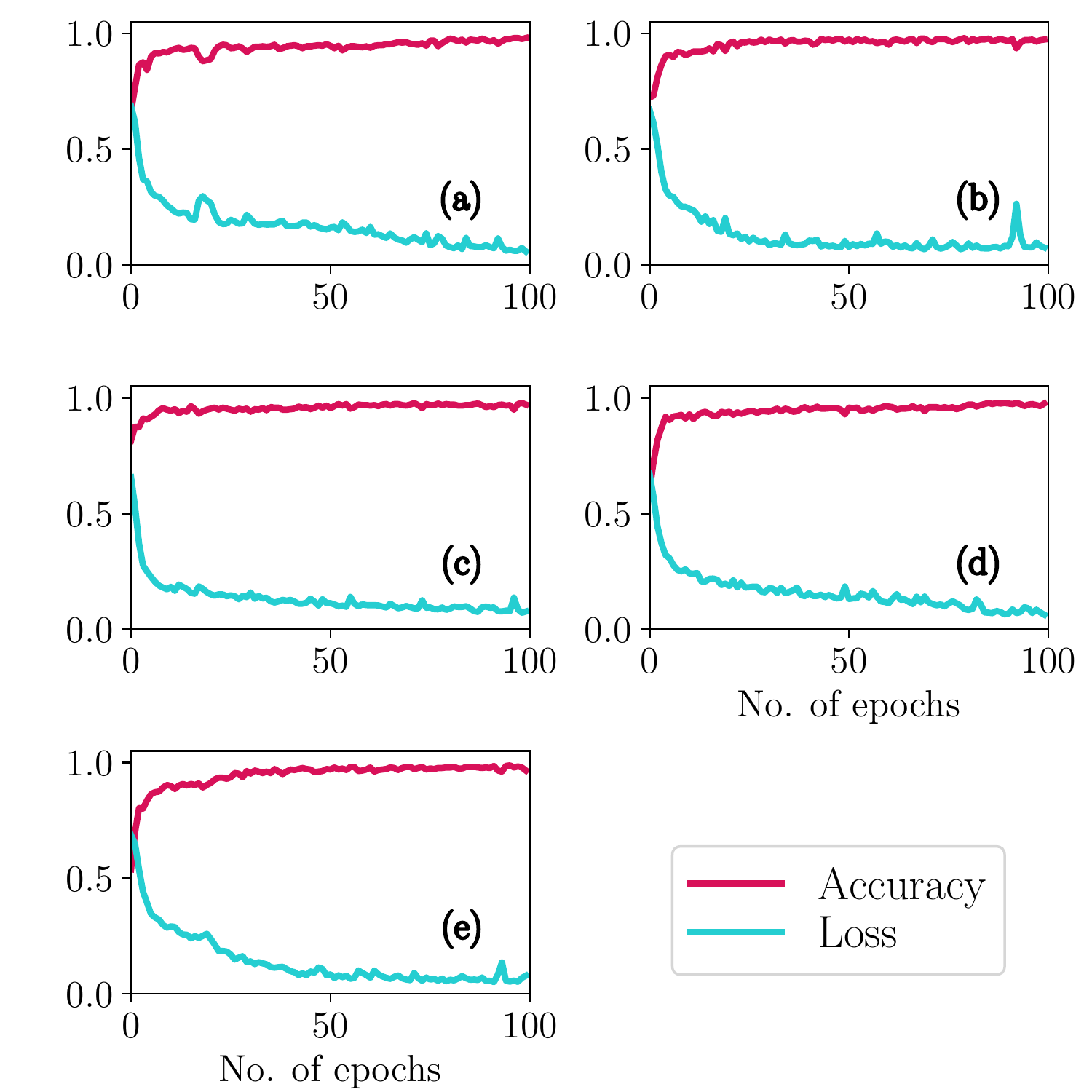}
	\caption{\label{fig:epoch} Plots showing the accuracy and loss values of the MLP model for the five different randomly shuffled data during the training process. While the number of epochs increases, the accuracy increases and loss value decreases}
\end{figure}

ANN is another well-known ML technique that replicates the working of biological neurons. Over the years, ANN has become a powerful tool for classification tasks. ANN has several layers and in each layer, there are certain number of nodes called neurons. An ANN which has more than one hidden layer is known as MLP \cite{goodfellow2016}. The schematic diagram of MLP is shown in Fig.~\ref{fig:dnn}. The considered network has 9 layers with one input layer, one output layer and 7 hidden layers. In our case, the number of neurons in the input layer is three which in turn represent the features ($f, \epsilon$ and $\delta$) of the considered system. We have only one neuron in the output layer which yields the output vector $\hat{Y}$  which can either be one or zero. These two numbers represent extreme and non-extreme events respectively. The number of neurons in the hidden layers are taken as (8, 16, 32, 32, 32, 16, 8). All the neurons in the MLP model are fully connected with each other. 
The interactions between the neurons in the subsequent layers are calculated by applying activation function $a$ to $z^{(l)}$, where $z^{(l)}$ takes the form
\begin{subequations}
	\begin{equation}
	\textbf{z}^{(l)} = \textbf{W}^{(l,l-1)}.\textbf{y}^{(l-1)}+\textbf{b}^{(l)},\label{bia}
	\end{equation}
	\begin{equation}
	\textbf{y}^{(l)} = a(\textbf{z}^{(l)}).
	\end{equation}
\end{subequations}
In the above equation \eqref{bia}, $\textbf{W}^{(l,l-1)}$ and $\textbf{b}^{(l)}$ represent the weight and bias matrices respectively. The notations $(l)$ and $(l-1)$ at the superscripts indicate respectively a layer and its previous layer in the network. Further, $\textbf{y}^{(l)}$ and $\textbf{y}^{(l-1)}$ respectively denote the output of the layers $l$ and $l-1$. The matrices $\textbf{W}$ and $\textbf{b}$ arbitrarily change during the training process in order to minimize the loss function $(L)$. In binary classification, the binary cross entropy is used as a loss function and is given by \cite{goodfellow2016}
\begin{equation}
L = -\frac{1}{N_{T}} \sum_{j=1}^{N_T} y_j.\log(p(y_j))+(1-y_j).\log(1-p(y_j)),
\end{equation}
where $N_T$ is the number of training examples, $y_j$ is the actual label of the $j^{th}$ data and $p(y_j)$ is the output for the $j^{th}$ data, that is probability of occurrence of extreme events for the given data. In all the hidden layers of the considered network, Rectified Linear Unit (ReLU) is used as an activation function and it is defined by
\begin{equation}
a_{ReLU}(z) = max(0,z).
\end{equation}
In the output layer, we use the sigmoid activation function  which is already pointed out in Eq.~\eqref{sigmoid_fn}. The working of ReLU and the sigmoid activation function can be visualized from Fig.~\ref{fig:ac_fn}. Adam optimizer \cite{kingma2014} is used to minimize the loss function $L$. The MLP model is trained with the training data for 100 epochs with a batch size equal to ten. The training process of MLP with five random data set are presented in Fig.~\ref{fig:epoch}. We can visualize from Fig.~\ref{fig:epoch} that when the number of epochs increases, the accuracy increases whereas the loss value decreases and these values settle at a particular value around epoch number 90.
\subsection{Comparison of ML models}
\par {In this section, we compare the working and training behaviour of the considered four ML models. As far as the common utility of them is concerned, the models LR and RF are used only for classification whereas the models SVM and MLP are used for both classification and regression. The basic functionality of each model is as follows, LR model has different decision boundaries with corresponding weights to classify the given data,  SVM seeks to find the best hyperplane in the feature space which separates the two classes, RF algorithm works on  the principle of ensembling the multiple decision trees and the MLP is working similar to that of the biological neural network and has multiple layers and nodes with corresponding weights and bias with some activation functions for making the decision.}
\par All the above four ML methods {are trained using the training set data. After training, they are allowed to predict the test data and the predicted labels are compared with the actual labels after fitting the training set data. In other words, after completion of the training process, the ML models are in the stage to predict whether the given new set of parameter values are corresponding to extreme or non-extreme.}
%------------------------------------------------------
\section{Performance of ML models}
\par To choose the best model among the four, for the prediction of occurrence of extreme events, we compare their performance {on the basis of the testing results}. The performance of these models are validated using the classification performance metrics namely, (a) confusion matrix, (b) accuracy, (c) precision, (d) recall and (e) F1-score of the models. The values of these metrics are given in Table~\ref{tab:result}. {In the following, we briefly discuss each one of the above mentioned performance metrics.} 
\begin{figure}[!ht]
	\centering
	\includegraphics[width=0.6\linewidth]{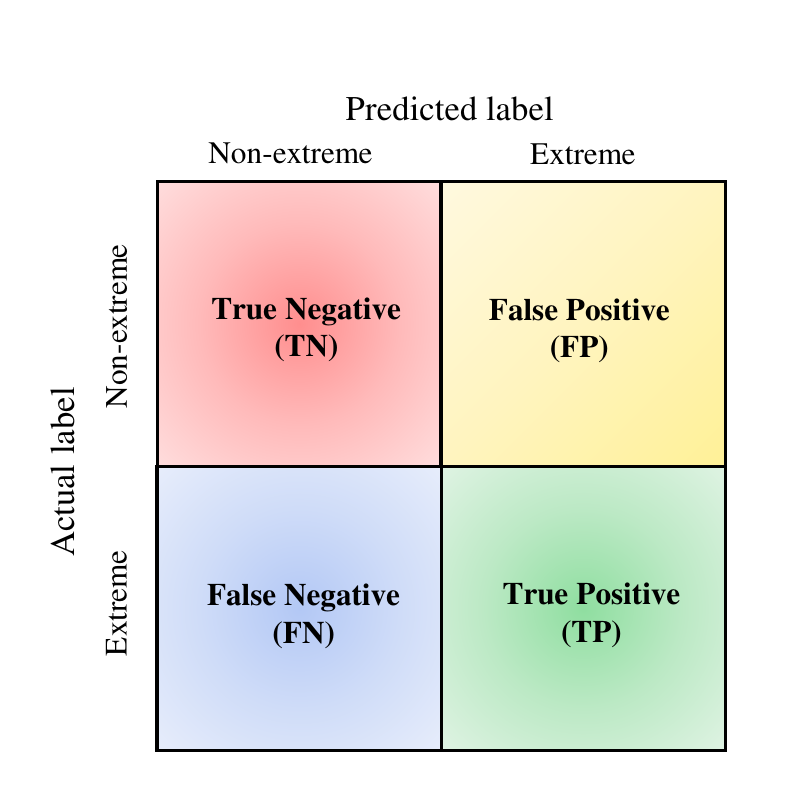}
	\caption{\label{fig:confusion} General representation of confusion matrix for the classification results of the ML  Models}
\end{figure}

\begin{enumerate}[label=\textbf{(\alph*)}]
	\item \textbf{Confusion matrix}
	\par A confusion matrix is an {$2 \times 2$} array {and the elements in that matrix denote the counts} of correct and incorrect predictions which help to understand how our classification models are confused to classify the given set of data. The pictorial representation of the confusion matrix is shown in Fig.~\ref{fig:confusion}. {From Fig.~{\ref{fig:confusion}}, we can see that the elements of the confusion matrix are denoting the number of True Positives (TP), True Negatives (TN), False Positives (FP) and False Negatives (FN). The definition of these metrics are taken as given below.}
	
	\begin{enumerate}[label=\textit{(\roman*)}]
		\item \textit{True Positive (TP)}
		\par {Let the actual label of the given input is extreme. If the ML model correctly predicts the output as extreme then that is known as True Positive.}
		
		\item \textit{True  Negative (TN)}
		\par {If the actual label is non-extreme and the outcome from the ML model is also non-extreme then it is True Negative.}
		
		\item \textit{False Positive (FP)}
		\par {The model predicts certain input as extreme but the actual label corresponding to that input is non-extreme. This type of prediction is defined as False Positive.}
		
		\item \textit{False Negative (FN)}
		\par {If the ML model wrongly predicts the label extreme as non-extreme, then it is called as False Negative.}
	\end{enumerate}
	
	\begin{figure}[!ht]
		\centering
		\includegraphics[width=1.0\linewidth]{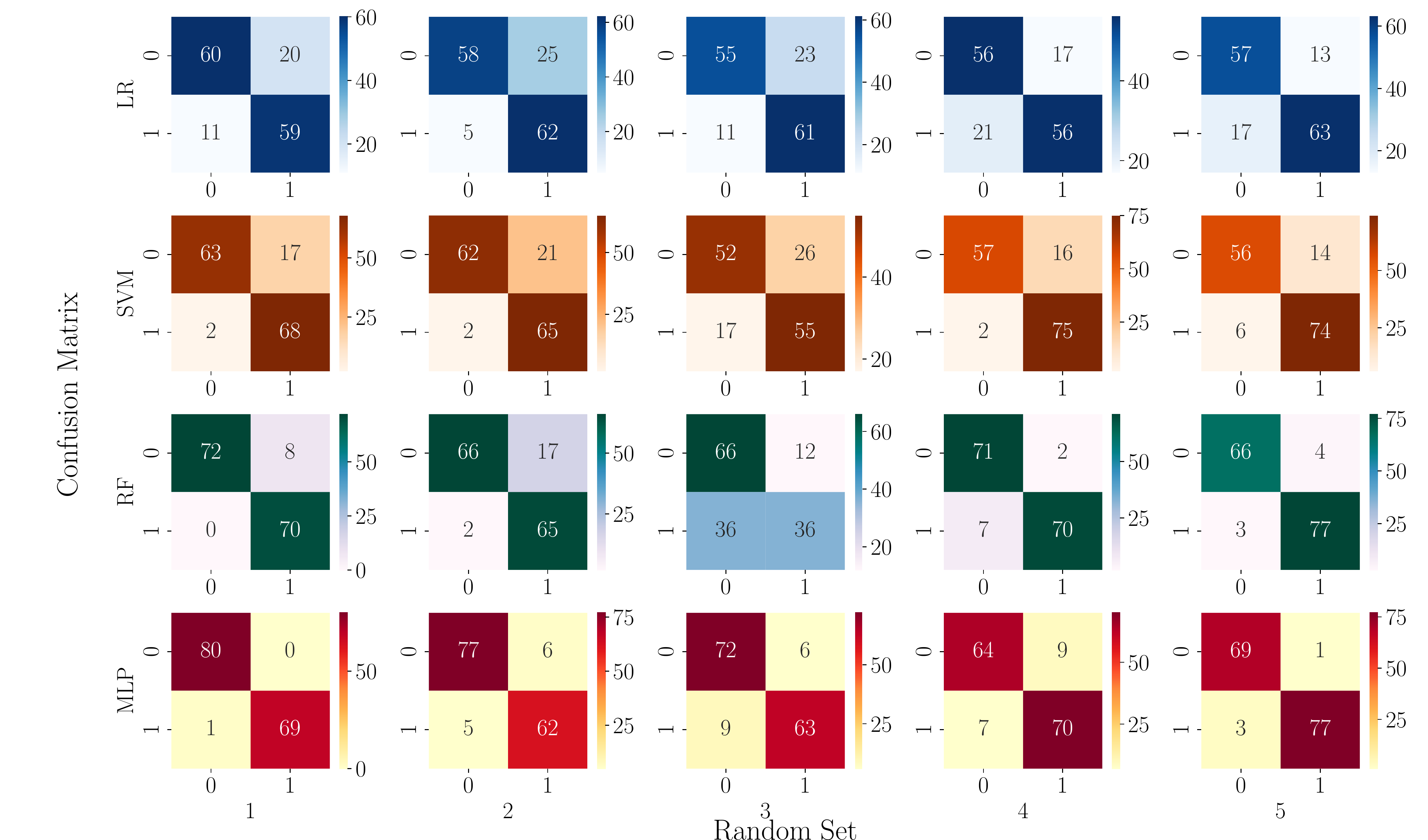}
		\caption{\label{fig:confusion_values} Representation of confusion matrix of the four ML models for five random data set}
	\end{figure}
	
	\par To capture the best model among the four we show the values of the confusion matrix in Fig.~\ref{fig:confusion_values}. Rows one to four respectively represent the confusion matrices of LR, SVM, RF and MLP.  Columns one to five correspond to the five randomly shuffled data sets. While examining the confusion matrix of the LR model we can see that the number of TN and TP are low while the number of FP and FN are high when compared with the other three models for all five data sets. In the SVM case, even though it gives acceptable TPs it also produces very low numbers for FNs for four sets of data. Since the count for the FPs is high, the SVM model wrongly predicts the result as extreme instead of non-extreme. Similarly, in the case of RF, the numbers of FPs and FNs are low except for the third data set where the numbers of FNs and TPs are very high. Finally, while inspecting the confusion matrices of the MLP model, we can find that the numbers of TPs and TNs are high whereas the numbers of FPs and FNs are very low for all the five data sets. In general, the model which has low numbers for FNs and FPs and high numbers for TPs and TFs is considered to be the best model for the classification task. Therefore in our case, MLP serves to be the best model while inspecting the confusion matrix alone.\\
	
	\item \textbf{Accuracy}
	\par The accuracy of the ML models is calculated by dividing the total number of correct classifications by the total number of classifications done \cite{chandramouli2018machine}, that is
	
	\begin{equation}
	\textrm{Accuracy} = \frac{\textrm{TP+TN}}{\textrm{TP+FP+FN+TN}}.\label{acc}
	\end{equation}
	
	{The numerator in equation {\eqref{acc}} is the sum of number of true positives and true negatives which inturn gives the total number of correct classifications done by the ML model.} Suppose one gets the value of the accuracy as 1.0 then it means that the considered ML model has correctly predicted both the extreme and non-extreme cases.
	
	\begin{figure}[!ht]
		\centering
		\includegraphics[width=0.6\linewidth]{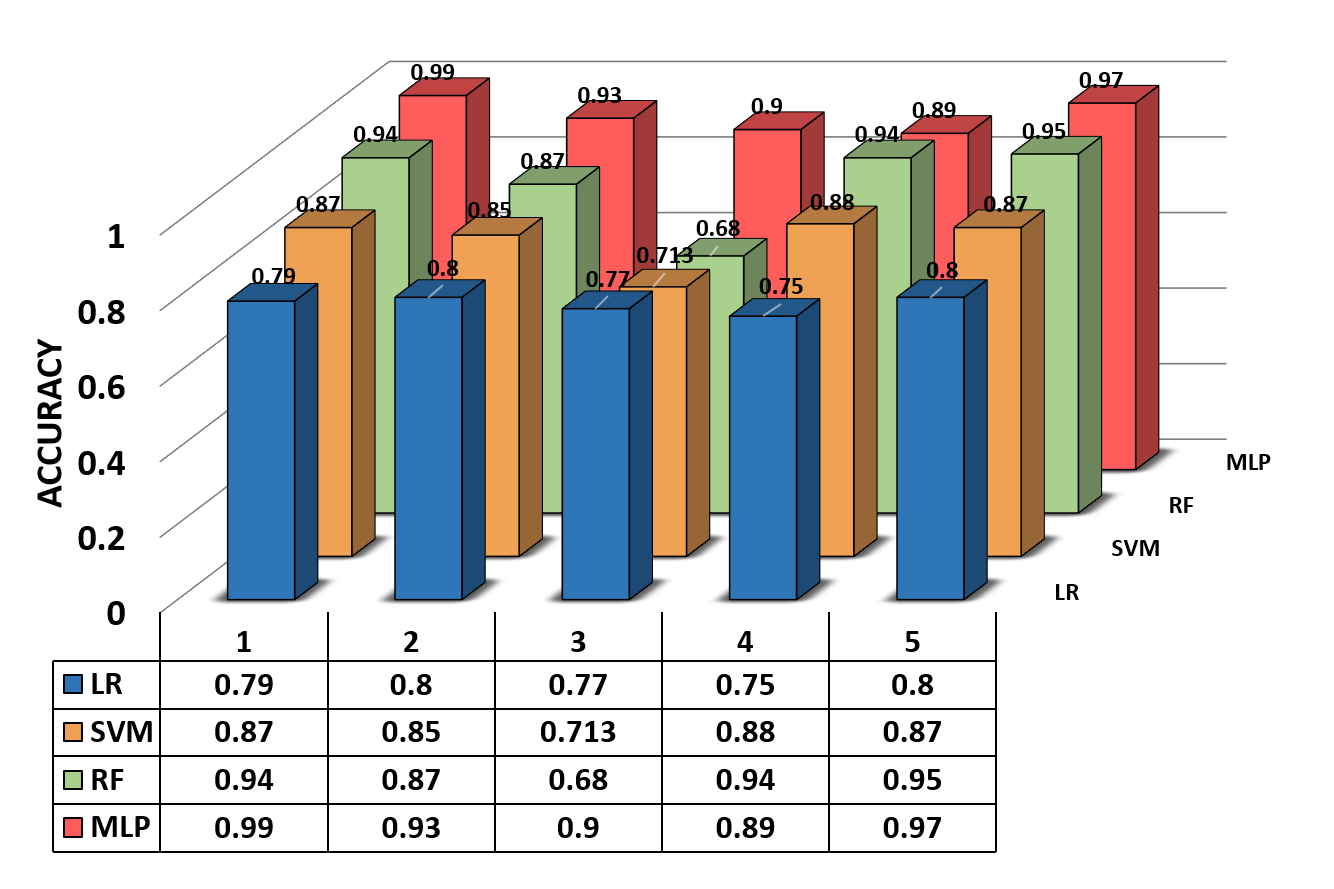}
		\caption{\label{fig:accuracy} Bar plot representating Accuracy of the ML models}
	\end{figure}
	\par The accuracy values of the four ML models that we have obtained during the testing process have been summarized as vertical bars in Fig.~\ref{fig:accuracy}. In the figure, colours blue, orange, green and red respectively represents the four ML models LR, SVM, RF and MLP. From the figure, one can notice that the LR model exhibits the range 0.75-0.8, the SVM model shows 0.7-0.88, for the RF model it is between 0.68-0.95 and for the MLP model the values lie between 0.89-0.98. It can be easily visualized from the figure that MLP has higher accuracy when compared with the other three models. Owing to high accuracy, MLP performs well in the considered classification task. \\
	
	\item \textbf{Precision}
	\par Precision denotes how often the model predicts the positive results (extreme cases) correctly. {In other words, precision is the ratio between the number of true positives and a total number of predicted positive outcomes.} This {can be} calculated using the formula \cite{chandramouli2018machine}
	\begin{equation}
	\textrm{Precision} = \frac{\textrm{TP}}{\textrm{TP+FP}}.
	\end{equation}
	If the precision is 1.0 then it means that the ML models' output has no false positives. 
	\begin{figure}[!ht]
		\centering
		\includegraphics[width=0.6\linewidth]{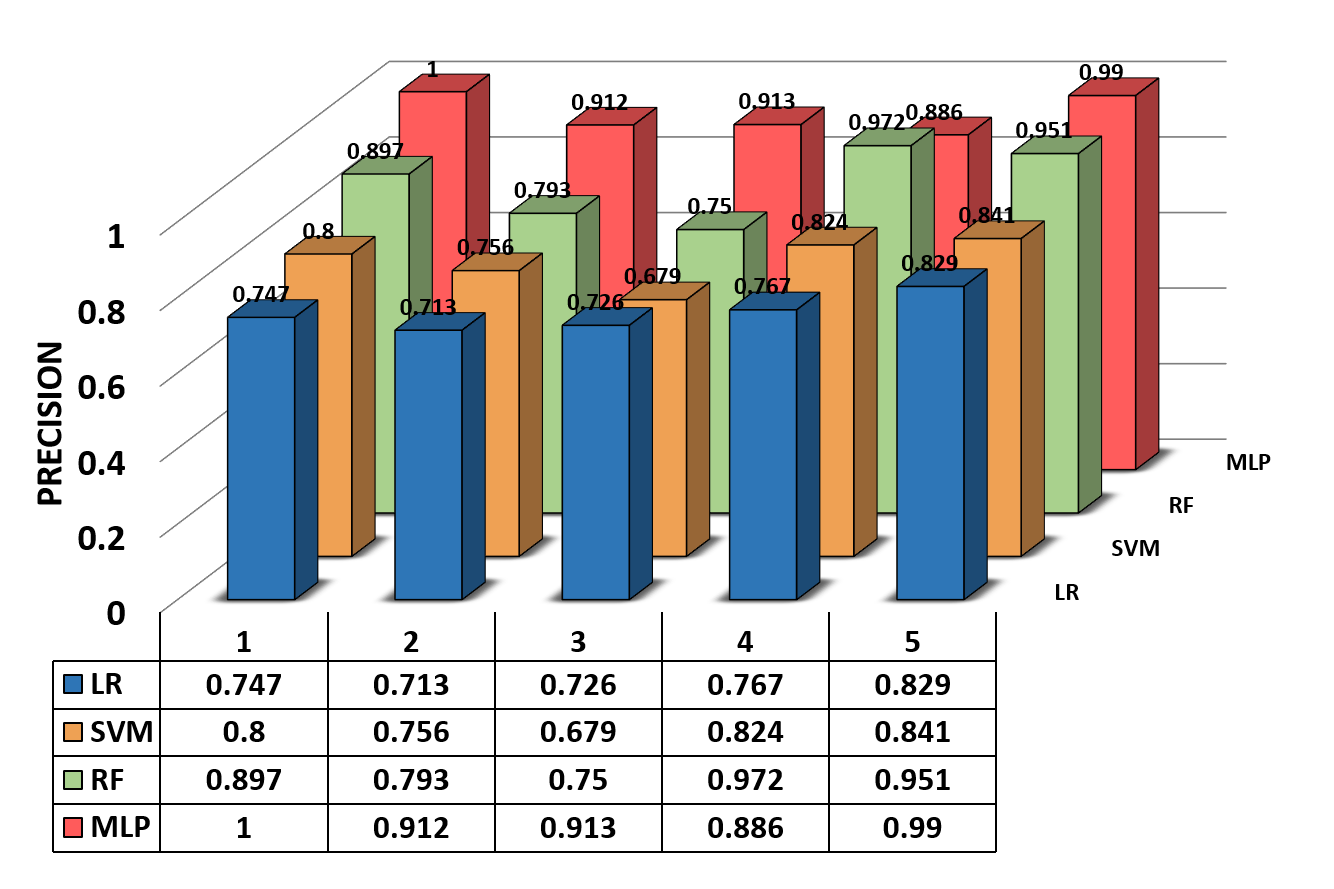}
		\caption{\label{fig:precision} Bar plot representating Precision of the ML models}
	\end{figure}
	\par The precision values of all the four ML models are plotted as bar plot in Fig.~\ref{fig:precision}. Examining the precision bars in Fig.~\ref{fig:precision} similar to that of the accuracy bars in Fig.~\ref{fig:accuracy} we observe that MLP has the highest precision metrics among the four models. It indicates that MLP outperforms well in predicting the positive results (in our case extreme events).
	
	\item \textbf{Recall}
	\par Recall indicates the proportion  of correct predictions of positives (extreme cases) to the total number of {actual} positive results \cite{chandramouli2018machine}, that is 
	\begin{equation}
	\textrm{Recall} = \frac{\textrm{TP}}{\textrm{TP+FN}}.
	\end{equation}
	
	If the recall is 1.0 then it means that the ML models' output has no false negatives. Since our foremost aim is to predict extreme events accurately in the considered system, we restrict ourselves in calculating Precision and Recall for positive results (extreme events case) alone.
	
	\begin{figure}[!ht]
		\centering
		\includegraphics[width=0.6\linewidth]{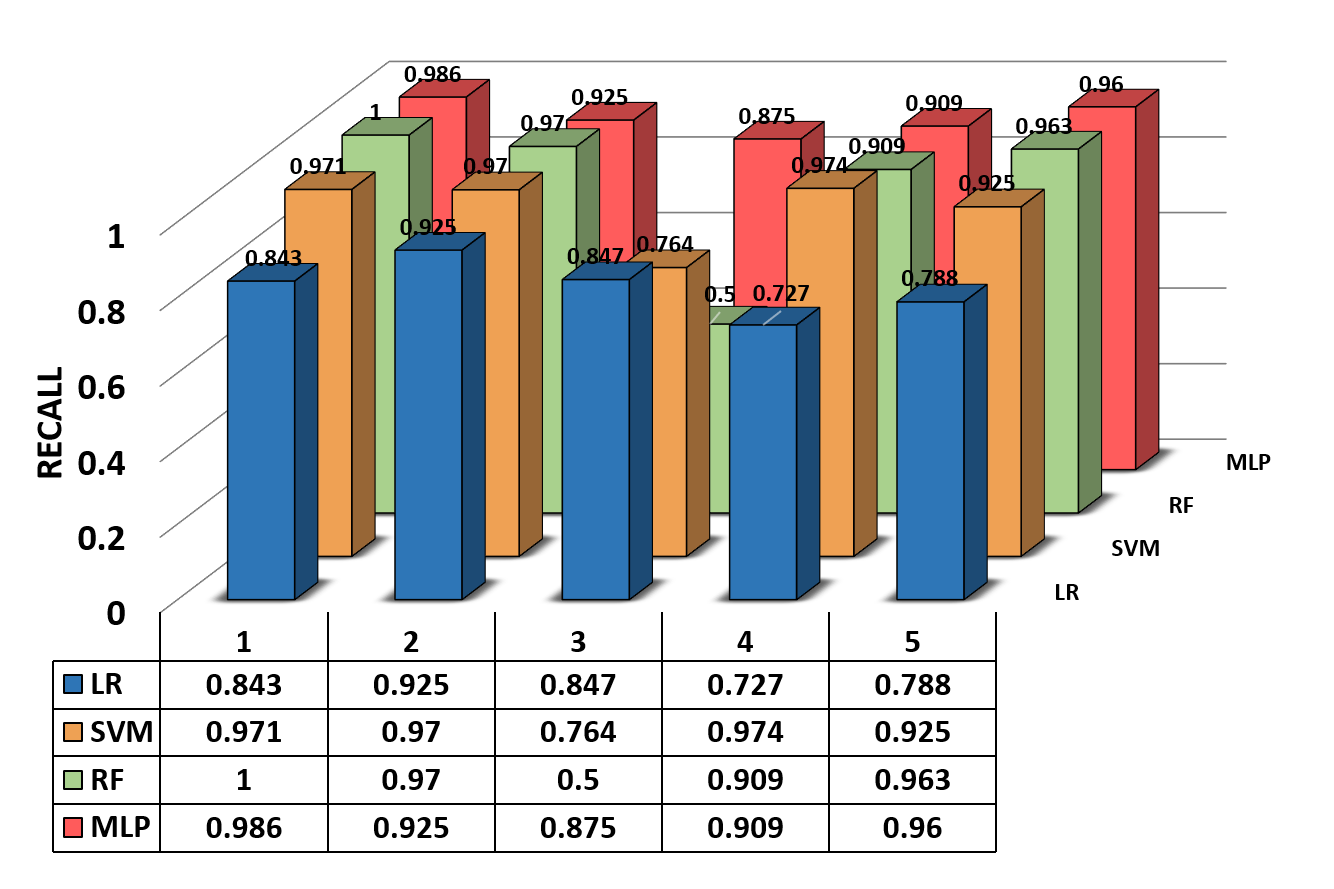}
		\caption{\label{fig:recall} Bar plot representating Recall of the ML models}
	\end{figure}
	\par The obtained recall values of the test results are plotted as bars in Fig.~\ref{fig:recall}. Upon analyzing the recall bars, we find that MLP has the highest recall values for all the five different data sets. This confirms that in the classification task, MLP predicts the outcome with the least number of FNs.

	\item \textbf{F1-score}
	\par Another measure that combines precision and recall is the F1 score. The F1 score is the harmonic mean of precision and recall. It is computed through the formula \cite{chandramouli2018machine}
	\begin{equation}
	\textrm{F1 score} = \frac{2\times\textrm{precision}\times \textrm{recall}}{\textrm{precision} + \textrm{recall}}.
	\end{equation}
	\begin{figure}[!ht]
		\centering
		\includegraphics[width=0.6\linewidth]{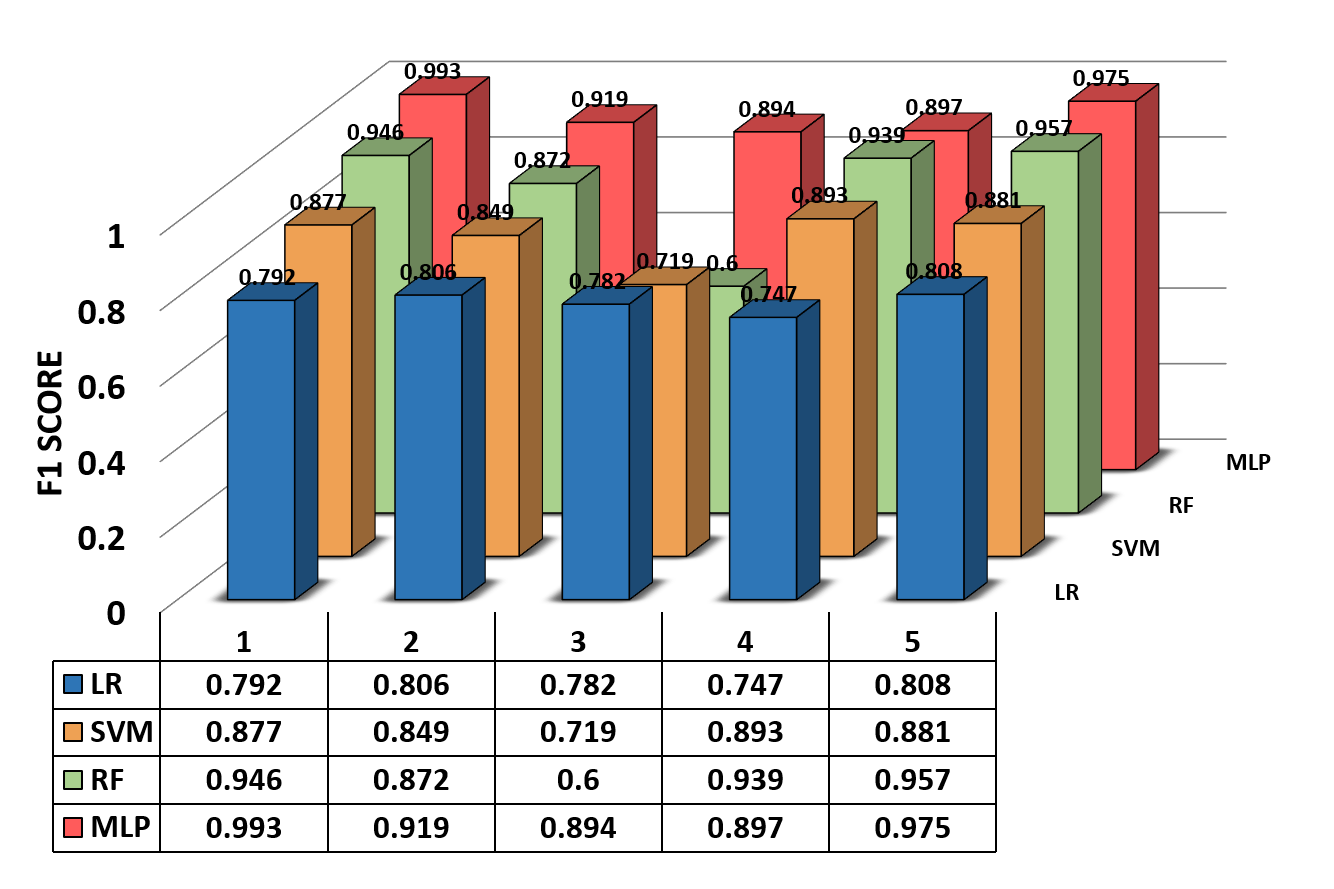}
		\caption{\label{fig:f1} Bar plot representating F1-Score of the ML models}
	\end{figure}
	\par In Fig.~\ref{fig:f1}, we plot the values of F1-Scores as a bar plot. The model which has a higher F1-Score has a higher balance between precision and recall. In our case, we find from Fig.~\ref{fig:f1} that MLP has the best balance between precision and recall implying that MLP classifies between extreme and non-extreme with higher values of TPs and lower values of FNs.

\end{enumerate}
\par All these results are completely consolidated in Table~\ref{tab:result}. One of our foremost aims is to determine a persistent ML method for the considered classification task. While inspecting the metrics of the MLP model, we find that the values of all metrics are better than the other three models. In particular, the MLP model has very low false positives and false negatives in the confusion matrices. These low values indicate that the created MLP model has significantly less chance to make wrong predictions. Even though the metrics of all other models have the best values for this particular set of data, the MLP model shows consistent performance in all five randomly shuffled data. In other words, the MLP model performs well in the prediction of both extreme and non-extreme cases and its outcome is robust to the given random set of parameter values.

\begin{table}[!ht]
	\scriptsize
	\centering
	\begin{tabular}{|c|c|c|c|c|c|c|c|c|c|c|}
		\hline
		\multicolumn{11}{|c|}{\textbf{Performance of the ML models}}\\\hline 
		&\multicolumn{5}{|c|}{\textbf{LR}}&\multicolumn{5}{|c|}{\textbf{SVM}}\\ \hline
		No.  & C. mat & Acc & Pre & Rec & F1 & C. mat & Acc & Pre & Rec & F1 \\ \hline
		1 & $\begin{matrix} 60 & 20 \\ 11 & 59 \end{matrix}$ & 0.79 & 0.747 & 0.843 & 0.792 & $\begin{matrix} 63 & 17\\2 & 68 \end{matrix}$ & 0.87 & 0.8 & 0.971 & 0.877 \\ \hline
		2 & $\begin{matrix} 58 & 25 \\ 5 & 62 \end{matrix}$ & 0.8 & 0.713 & 0.925 & 0.806 & $\begin{matrix} 62 & 21\\2 & 65 \end{matrix}$ & 0.85 & 0.756 & 0.97 & 0.849 \\ \hline
		3 & $\begin{matrix} 55 & 23 \\ 11 & 61 \end{matrix}$ & 0.77 & 0.726 & 0.847 & 0.782 & $\begin{matrix} 52 & 26\\17 & 55 \end{matrix}$ & 0.713 & 0.679 & 0.764 & 0.719 \\ \hline
		4 & $\begin{matrix} 56 & 17 \\ 21 & 56 \end{matrix}$ & 0.75 & 0.767 & 0.727 & 0.747 & $\begin{matrix} 57 & 16\\2 & 75 \end{matrix}$ & 0.88 & 0.824 & 0.974 & 0.893 \\ \hline
		5 & $\begin{matrix} 57 & 13 \\ 17 & 63 \end{matrix}$ & 0.8 & 0.829 & 0.788 & 0.808 & $\begin{matrix} 56 & 14\\6 & 74 \end{matrix}$ & 0.87 & 0.841 & 0.925 & 0.881 \\ \hline\hline\hline
		&\multicolumn{5}{|c|}{\textbf{RF}}&\multicolumn{5}{|c|}{\textbf{MLP}}\\ \hline
		No.  & C. mat & Acc & Pre & Rec & F1 & C. mat & Acc & Pre & Rec & F1 \\ \hline
		1  & $\begin{matrix} 72 & 8 \\ 0 & 70 \end{matrix}$ & 0.94 & 0.897 & 1.0 & 0.946 & $\begin{matrix} 80 & 0\\1 & 69 \end{matrix}$ & 0.99 & 1.0 & 0.986 & 0.993\\ \hline
		2  & $\begin{matrix} 66 & 17 \\ 2 & 65 \end{matrix}$ & 0.87 & 0.793 & 0.97 & 0.872 & $\begin{matrix} 77 & 6\\5 & 62 \end{matrix}$ & 0.93 & 0.912 & 0.925 & 0.919\\ \hline
		3  & $\begin{matrix} 66 & 12 \\ 36 & 36 \end{matrix}$ & 0.68 & 0.75 & 0.5 & 0.6 & $\begin{matrix} 72 & 6\\9 & 63 \end{matrix}$ & 0.9 & 0.913 & 0.875 & 0.894\\ \hline
		4  & $\begin{matrix} 71 & 2 \\ 7 & 70 \end{matrix}$ & 0.94 & 0.972 & 0.909 & 0.939 & $\begin{matrix} 64 & 9\\7 & 70 \end{matrix}$ & 0.89 & 0.886 & 0.909 & 0.897\\ \hline
		5  & $\begin{matrix} 66 & 4 \\ 3 & 77 \end{matrix}$ & 0.95 & 0.951 & 0.963 & 0.957 & $\begin{matrix} 69 & 1\\3 & 77 \end{matrix}$ & 0.97 & 0.99 & 0.96 & 0.975\\ \hline
	\end{tabular}
	\caption{Table for the performance metrics of the ML models. Here C. mat, Acc, Pre, Rec and F1 represent the confusion matrix, accuracy, precision, recall and F1-score respectively}
	\label{tab:result}
\end{table}

\section{Conclusion}
\par In this work, we have considered a non-polynomial mechanical system with external forcing, parametric drive and time delayed feedback and examined the prediction of the occurrence of extreme events with the help of ML tools. We set the parameters, namely external forcing ($f$), the strength of the parametric drive ($\epsilon$) and the strength of the time delayed feedback ($\delta$) as input of the ML models. The data for prediction is taken in six different combinations of the three parameter values. We have also randomly shuffled the data and split them into two categories, namely training set and testing set. We have employed four ML models, namely LR, SVM, RF and MLP. We have trained each model with the five random set data and then predicted the occurrence of extreme events by giving test data set into the considered ML models. We set the output as either one or zero denoting the occurrence of extreme events or absence of extreme events in \eqref{gen}. We have evaluated the performance of each model based on the confusion matrix, accuracy, precision, recall and F1 score of each model. The performance metrics are shown using bar plots. On examining the performance of each model, we found that the performance of the MLP model is high when compared to the other three. Also, the performance of MLP turns out to be consistent in all five random data set. From the outcome, we conclude that for the considered non-polynomial mechanical system, as far as the prediction of the emergence of extreme events from the parameter values is concerned, one can optimally use the MLP model. {It gives us the answer to the question: \textit{Is there any occurrence of extreme events in the system for the given input parameter values?}} with high accuracy. Once the parameter values corresponding to the extreme regime are predicted, we can deploy precaution measures to prevent the emergence of such events and formulate the safeguarding strategies. The methodology adopted in this paper can further be extended to other physical, mechanical and biological systems.\\

\section*{Acknowledgements}
JM thanks RUSA 2.0 project for providing a fellowship to carry out this work. SS  thanks  to the  Department  of  Science  and  Technology (DST), Government of India, for support through INSPIRE Fellowship (IF170319). The work of AV forms a part of a research project sponsored by DST under Grant No. EMR/2017/002813. MS acknowledges the Department of Science and Technology (DST) under PURSE Phase-II program for providing financial support in procuring high-performance computer which highly assisted this work.
\section*{Authors Contribution Statement}
All the authors contributed equally to the preparation of this manuscript.
\section*{Data Availability Statement}
The data that support the findings of this study are available
within the article.

\bibliography{mybibfile}
\end{document}